\documentclass[12]{article}
\usepackage{hyperref}
\usepackage{amsmath}
\usepackage{graphicx}
\usepackage{amsfonts}
\usepackage{amssymb}
\usepackage{algorithm}
\usepackage{algorithmic}

\begin{document}

\title{Discriminative Metric Learning with Deep Forest}
\author{Lev V. Utkin$^{1}$ and Mikhail A. Ryabinin$^{2}$\\{\normalsize Department of Telematics}\\{\normalsize Peter the Great St.Petersburg Polytechnic University}\\{\normalsize St.Petersburg, Russia}\\{\normalsize e-mail: }$^{1}${\normalsize lev.utkin@gmail.com, }$^{2}%
${\normalsize mihail-ryabinin@yandex.ru}}
\date{}
\maketitle

\begin{abstract}
A Discriminative Deep Forest (DisDF) as a metric learning algorithm is
proposed in the paper. It is based on the Deep Forest or gcForest proposed by
Zhou and Feng and can be viewed as a gcForest modification. The case of the
fully supervised learning is studied when the class labels of individual
training examples are known. The main idea underlying the algorithm is to
assign weights to decision trees in random forest in order to reduce distances
between objects from the same class and to increase them between objects from
different classes. The weights are training parameters. A specific objective
function which combines Euclidean and Manhattan distances and simplifies the
optimization problem for training the DisDF is proposed. The numerical
experiments illustrate the proposed distance metric algorithm.


\end{abstract}

\section{Introduction}

Real-world data usually has a high-dimensionality and a non-linear complex
structure. In order to improve the performance of many classification
algorithms especially those relying on distance computations (nearest neighbor
classifiers, support vector machines, etc.) the distance metric learning
methods are widely applied. Bellet et al. \cite{Bellet-etal-2013} selected
three main groups of distance metric learning methods, which are defined by
conditions of the available information about class labels of training
examples. The first group consists of fully supervised algorithms for which we
have a set of labeled training examples $S=\{(\mathbf{x}_{i},y_{i}%
),\ i=1,...,n\}$ such that every example has a label of a class $y_{i}%
\in \mathcal{Y}$, i.e., the class labels of individual training examples are
known The second group consists of weakly supervised algorithms. This group is
characterized by the lack of class labels for training every example, but
there is a side information in the form of constraints corresponding to the
semantic similarity or dissimilarity of pairs of training examples. This can
be seen as having label information only at the pair level. The third group
consists of semi-supervised algorithms for which there is a part of data that
are labeled or belong to similarity constraints, but another part consists of
fully unlabeled data.

We study the first group of algorithms in this paper. The main goal of fully
supervised distance metric learning is to use discriminative information to
keep all the data samples in the same class close and those from different
classes separated \cite{Mu-Ding-2013}. As indicated by Yang and Jin
\cite{Yang-Jin-2006}, unlike most supervised learning algorithms where
training examples are given class labels, the training examples of supervised
distance metric learning is cast into pairwise constraints: the equivalence
constraints where pairs of data points that belong to the same classes, and
inequivalence constraints where pairs of data points belong to different classes.

Metric learning approaches were reviewed in
\cite{Bellet-etal-2013,LeCapitaine-2016,Kulis-2012,Zheng-etal-2016}. The basic
idea underlying the metric learning solution is that the distance between
similar objects should be smaller than the distance between different objects.
If we have two observation vectors $\mathbf{x}_{i}\in \mathbb{R}^{m}$ and
$\mathbf{x}_{j}\in \mathbb{R}^{m}$ from a training set, and the similarity of
objects is defined by their belonging to the same class, then the distance
$d(\mathbf{x}_{i},\mathbf{x}_{j})$ between the vectors should be minimized if
$\mathbf{x}_{i}$ and $\mathbf{x}_{j}$ belong to the same class, and it should
be maximized if $\mathbf{x}_{i}$ and $\mathbf{x}_{j}$ are from different
classes. Several review papers analyze various methods and algorithms of
metric learning
\cite{Kedem-etal-2012,Norouzi-etal-2012,Xu-Weinberger-Chapelle-2012}. A
powerful implementation of the metric learning dealing with non-linear data
structures is the so-called Siamese neural network introduced by Bromley et
al. \cite{Bromley-etal-1993} in order to solve signature verification as a
problem of image matching. This network consists of two identical sub-networks
joined at their outputs. The two sub-networks extract features from two input
examples during training, while the joining neuron measures the distance
between the two feature vectors. The Siamese architecture has been exploited
in many applications, for example, in face verification
\cite{Chopra-etal-2005}, in the one-shot learning in which predictions are
made given only a single example of each new class \cite{Koch-etal-2015}, in
constructing an inertial gesture classification \cite{Berlemont-etal-2015}, in
deep learning \cite{Wang-etal-2016}, in extracting speaker-specific
information \cite{Chen-Salman-2011}, for face verification in the wild
\cite{Hu-Lu-Tan-2014}. This is only a part of successful applications of
Siamese neural networks. Many modifications of Siamese networks have been
developed, including fully-convolutional Siamese networks
\cite{Bertinetto-etal-2016}, Siamese networks combined with a gradient
boosting classifier \cite{Leal-Taixe-etal-2016}, Siamese networks with the
triangular similarity metric \cite{Zheng-etal-2016}.

A new powerful method, which can be viewed as an alternative to deep
neural networks, is the deep forest proposed by Zhou and Feng
\cite{Zhou-Feng-2017} and called the gcForest. It can be compared with a
multi-layer neural network structure, but each layer in the gcForest contains
many random forests instead of neurons. The gsForest can be regarded as an
multi-layer ensemble of decision tree ensembles. Zhou and Feng
\cite{Zhou-Feng-2017} point out that their approach is highly competitive to
deep neural networks. In contrast to deep neural networks which require great
effort in hyperparameter tuning and large-scale training data, gcForest is
much easier to train and can perfectly work when there are only small-scale
training data. The deep forest solves the tasks of classification and
regression. Therefore, by taking into account its advantages, it is important
to modify it in order to develop a structure solving the metric learning task.
We propose the so-called Discriminative Deep Forest (DisDF) which is a
discriminative distance metric learning algorithm. It is based on the gcForest
proposed by Zhou and Feng \cite{Zhou-Feng-2017} and can be viewed as its
modification. The main idea underlying the proposed DisDF is to assign weights
to decision trees in random forest in order reduce distances between pairs of
examples from the same class and to increase them between pairs of examples
from different classes. We define the class distributions in the deep forest
as the weighted sum of the tree class probabilities where the weights are
determined by solving an optimization problem with the contrastive loss
function as an objective function. The weights are viewed as training
parameters. We also apply the greedy algorithm for training the DisDF, i.e.,
the weights are successively computed for every layer or level of the forest
cascade. In order to efficiently find optimal weights, we propose to modify
the standard contrastive loss in a way that makes the loss function to be
convex with respect to the weights. This modification is carried out by
combining Euclidean and Manhattan distances in the loss function. Moreover, we
reduce the optimization problem for computing the weights to the standard
convex quadratic optimization problem with linear constraints whose solution
does not meet difficulties. For large-scale data, we apply the well-known
Frank-Wolfe algorithm \cite{Frank-Wolfe-1956} which is very simple when the
feasible set of weights is the unit simplex.

The paper can be viewed as an extension of the results obtained by Utkin and
Ryabinin \cite{Utkin-Ryabinin-2017} where the Siamese Deep Forest has been
proposed. The main difference of the presented paper from
\cite{Utkin-Ryabinin-2017} is that the case of the weakly supervised learning
was used in the Siamese Deep Forest when there are no information about the
class labels of individual training examples, but only information in the form
of sets of semantically similar pairs is available. Now we study the case of
the fully supervised learning when the class labels of individual training
examples are known.

The paper is organized as follows. Section 2 gives a very short introduction
into the gcForest proposed by Zhou and Feng \cite{Zhou-Feng-2017}. The idea to
assign weights to trees in random forests, which allows us to construct the
DisDF, is considered in Section 3 in detail. Algorithms for training and
testing the DisDF are considered in Section 4. Section 5 provides algorithms
for dealing with large-scale training data. Numerical experiments with real
data illustrating the proposed DisDF are given in Section 6. Concluding
remarks are provided in Section 7.

\section{Deep Forest}

According to \cite{Zhou-Feng-2017}, the gcForest generates a deep forest
ensemble, with a cascade structure. Representation learning in deep neural
networks mostly relies on the layer-by-layer processing of raw features. The
gcForest representational learning ability can be further enhanced by the
so-called multi-grained scanning. Each level of cascade structure receives
feature information processed by its preceding level, and outputs its
processing result to the next level. Moreover, each cascade level is an
ensemble of decision tree forests. We do not consider in detail the
Multi-Grained Scanning where sliding windows are used to scan the raw features
because this part of the deep forest is the same in the DisDF. However, the
most interesting component of the gcForest from the DisDF construction point
of view is the cascade forest.%

%TCIMACRO{\FRAME{ftbpFU}{4.1537in}{1.8502in}{0pt}{\Qcb{The architecture of the
%cascade forest \cite{Zhou-Feng-2017}}}{\Qlb{fig:cascade_forest}}%
%{forest_cascade.png}{\special{ language "Scientific Word";  type "GRAPHIC";
%maintain-aspect-ratio TRUE;  display "USEDEF";  valid_file "F";
%width 4.1537in;  height 1.8502in;  depth 0pt;  original-width 9.3292in;
%original-height 4.1397in;  cropleft "0";  croptop "1";  cropright "1";
%cropbottom "0";  filename 'Forest_Cascade.png';file-properties "XNPEU";}} }%
%BeginExpansion
\begin{figure}
[ptb]
\begin{center}
\includegraphics[
%natheight=4.139700in,
%natwidth=9.329200in,
height=1.8502in,
width=4.1537in
]%
{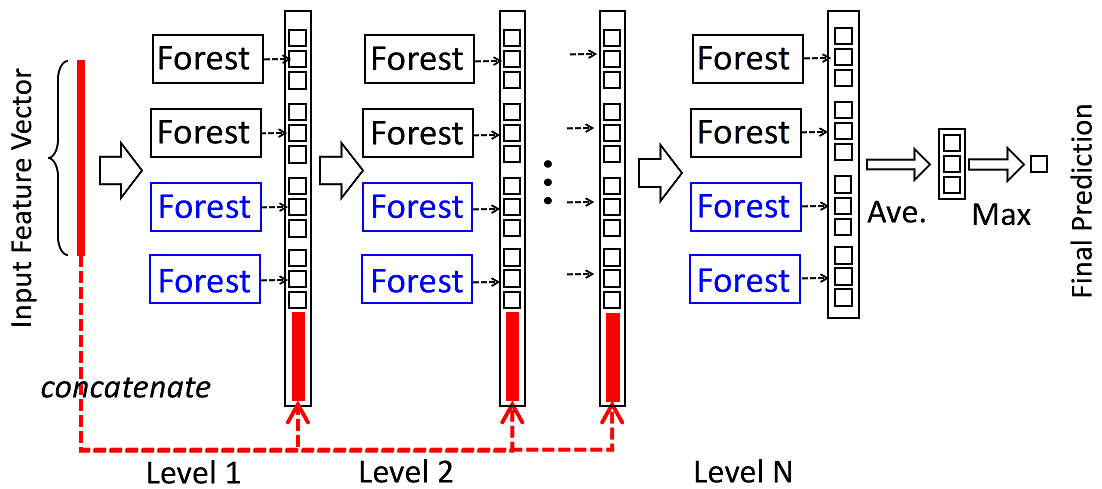}%
\caption{The architecture of the cascade forest \cite{Zhou-Feng-2017}}%
\label{fig:cascade_forest}%
\end{center}
\end{figure}
%EndExpansion

Given an instance, each forest produces an estimate of class distribution by
counting the percentage of different classes of examples at the leaf node
where the concerned instance falls into, and then averaging across all trees
in the same forest. The class distribution forms a class vector, which is then
concatenated with the original vector to be input to the next level of
cascade. The usage of the class vector as a result of the random forest
classification is very similar to the idea underlying the stacking method
\cite{Wolpert-1992}. The stacking algorithm trains the first-level learners
using the original training data set. Then it generates a new data set for
training the second-level learner (meta-learner) such that the outputs of the
first-level learners are regarded as input features for the second-level
learner while the original labels are still regarded as labels of the new
training data. In fact, the class vectors in the gcForest can be viewed as the
meta-learners. In contrast to the stacking algorithm, the gcForest
simultaneously uses the original vector and the class vectors (meta-learners)
at the next level of cascade by means of their concatenation. This implies
that the feature vector is enlarged and enlarged after every cascade level.
The architecture of the cascade proposed by Zhou and Feng
\cite{Zhou-Feng-2017} is shown in Fig. \ref{fig:cascade_forest}. It can be
seen from the figure that each level of the cascade consists of two different
pairs of random forests which generate 3-dimensional class vectors
concatenated each other and with the original input. After the last level, we
have the feature representation of the input feature vector, which can be
classified in order to get the final prediction. Zhou and Feng
\cite{Zhou-Feng-2017} propose to use different forests at every level in order
to provide the diversity which is an important requirement for the random
forest construction.

\section{Weighted averages in forests}

The DisDF aims to provide large distances between pairs of vectors belonging
to the same class and small distances between vectors from different classes.

In order to achieve the above aim, we modify ideas provided by Xiong et al.
\cite{Xiong-etal-2012} and Dong et al. \cite{Dong-Du-Zhang-2015}. Xiong et al.
\cite{Xiong-etal-2012} considered an algorithm for solving the metric learning
problem by means of the random forests. The proposed metric is able to
implicitly adapt its distance function throughout the feature space. Dong et
al. \cite{Dong-Du-Zhang-2015} proposed a random forest metric learning
algorithm which combines semi-multiple metrics with random forests to better
separate the desired targets and background in detecting and identifying
target pixels based on specific spectral signatures in hyperspectral image
processing. A common idea underlying the metric learning algorithms in
\cite{Dong-Du-Zhang-2015} and \cite{Xiong-etal-2012} is to define a distance
measure between a pair of training elements $\mathbf{x}_{i}$ and
$\mathbf{x}_{j}$ for a combination of trees as average of some special
functions of the training elements. For example, if a random forest is a
combination of $T$ decision trees $\{f_{t}(\mathbf{x}),t=1,...,T\}$, then the
distance measure is
\begin{equation}
d(\mathbf{x}_{i},\mathbf{x}_{j})=T^{-1}\sum_{t=1}^{T}f_{t}(\psi(\mathbf{x}%
_{i},\mathbf{x}_{j})). \label{DisDF_30}%
\end{equation}
Here $\psi(\mathbf{x}_{i},\mathbf{x}_{j})$ is a function specifically defined
in \cite{Dong-Du-Zhang-2015} and \cite{Xiong-etal-2012}.

The idea of the distance measure (\ref{DisDF_30}) produced by a random forest
combined with the idea of probability distributions of classes for producing
new augmented feature vectors after every level of the cascade forest proposed
by Zhou and Feng \cite{Zhou-Feng-2017} can be a basis for the modification of
the gcForest which produce a new feature representation for efficient metric
learning. According to \cite{Zhou-Feng-2017}, each forest of a cascade level
produces an estimate of the class probability distribution by counting the
percentage of different classes of training examples at a leaf node where the
concerned instance falls into, and then averaging across all trees in the same
forest. In contrast to this approach for computing the class probability
distribution, we propose to define the class distribution as a weighted sum of
the tree class probabilities. The weights can be viewed as training
parameters. They are optimized in order to reduce distances between examples
of the same class and to increase them between examples from different classes.

We apply the greedy algorithm for training the DisDF, i.e., we train
separately every level starting from the first level such that every next
level uses results of training obtained at the previous level.

Let us introduce notations for indices corresponding to different deep forest
components. The indices and their sets of values are shown in Table
\ref{t:DisDF_1}. One can see from Table \ref{t:DisDF_1}, that there are $Q$
levels of the deep forest, every level contains $M_{q}$ forests such that
every forest consists of $T_{k,q}$ trees. It is supposed that all training
examples are divided into $C$ classes.%

%TCIMACRO{\TeXButton{B}{\begin{table}[tbp] \centering}}%
%BeginExpansion
\begin{table}[tbp] \centering
%EndExpansion
\caption{Notations for indices}%
\begin{tabular}
[c]{cc}\hline
type & index\\ \hline
cascade level & $q=1,...,Q$\\ \hline
forest & $k=1,...,M_{q}$\\ \hline
tree & $t=1,...,T_{k,q}$\\ \hline
class & $c=1,...,C$\\ \hline
\end{tabular}
\label{t:DisDF_1}%
%TCIMACRO{\TeXButton{E}{\end{table}}}%
%BeginExpansion
\end{table}%
%EndExpansion

Suppose we have trained trees in the DisDF. According to \cite{Zhou-Feng-2017}%
, the class distribution forms a class vector which is then concatenated with
the original vector to be input to the next level of cascade. Suppose an
origin vector is $\mathbf{x}_{i}$, and the $p_{i,c}^{(t,k,q)}$ is the
probability of class $c$ for $\mathbf{x}_{i}$ produced by the $t$-th tree from
the $k$-th forest at the cascade level $q$. Below we use the triple index
$(t,k,q)$ in order to indicate that the element belongs to the $t$-th tree
from the $k$-th forest at the cascade level $q$. Following the results given
in \cite{Zhou-Feng-2017}, the element $v_{i,c}^{(k,q)}$ of the class vector
corresponding to class $c$ and produced by the $k$-th forest in the gcForest
is determined as%
\begin{equation}
v_{i,c}^{(k,q)}=T_{k,q}^{-1}\sum_{t=1}^{T_{k,q}}p_{i,c}^{(t,k,q)}.
\label{DisDF_36}%
\end{equation}

Denote the obtained class vector as $\mathbf{v}_{i}^{(k,q)}=(v_{i,1}%
^{(k,q)},...,v_{i,C}^{(k,q)})$ and the concatenated $M_{q}$ vectors
$\mathbf{v}_{i}^{(k,q)}$ as $\mathbf{v}_{i}^{(q)}$. Then the concatenated
vector $\mathbf{x}_{i}^{(1)}$ after the first level of the cascade is%
\[
\mathbf{x}_{i}^{(1)}=\left(  \mathbf{x}_{i},\mathbf{v}_{i}^{(1,1)}%
,....,\mathbf{v}_{i}^{(M_{1},1)}\right)  =\left(  \mathbf{x}_{i}%
,\mathbf{v}_{i}^{(k,1)},k=1,...,M_{1}\right)  =\left(  \mathbf{x}%
_{i},\mathbf{v}_{i}^{(1)}\right)  .
\]
It is composed of the original vector $\mathbf{x}_{i}$ and $M_{1}$ class
vectors obtained from $M_{1}$ forests at the first level. In the same way, we
can write the concatenated vector $\mathbf{x}_{i}^{(q)}$ after the $q$-th
level of the cascade as%
\begin{align}
\mathbf{x}_{i}^{(q)}  &  =\left(  \mathbf{x}_{i}^{(q-1)},\mathbf{v}%
_{i}^{(1,q)},....,\mathbf{v}_{i}^{(M_{q},q)}\right) \nonumber \\
&  =\left(  \mathbf{x}_{i}^{(q-1)},\mathbf{v}_{i}^{(k,q)},\ k=1,...,M_{q}%
\right)  =\left(  \mathbf{x}_{i}^{(q-1)},\mathbf{v}_{i}^{(q)}\right)  .
\label{DisDF_40}%
\end{align}

Below we omit the index $q$ in order to reduce the number of indices because
all derivations will concern only level $q$, where $q$ may be arbitrary from
$1$ to $Q$.

\begin{figure}
[ptb]
\begin{center}
\includegraphics[
%natheight=2.364500in,
%natwidth=9.131600in,
height=1.3602in,
width=5.2041in
]%
{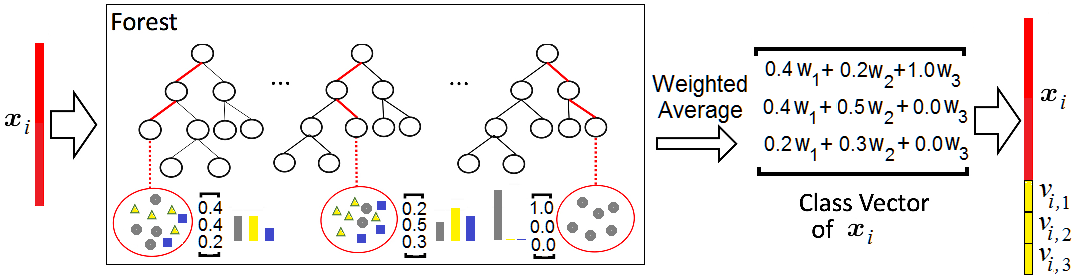}%
\caption{An illustration of the class vector generation taking into account
the weights}%
\label{fig:weighted_class}%
\end{center}
\end{figure}

The vector $\mathbf{x}_{i}$ in (\ref{DisDF_40}) is derived in accordance with
the gcForest algorithm \cite{Zhou-Feng-2017} by using (\ref{DisDF_36}). We
propose to change the method for computing elements $v_{i,c}^{(k)}$ of the
class vector in the DisDF, namely, the averaging (\ref{DisDF_36}) is replaced
with the weighted sum of the form:
\begin{equation}
v_{i,c}^{(k)}=\sum_{t=1}^{T_{k}}p_{i,c}^{(t,k)}w^{(t,k)}. \label{DisDF_41}%
\end{equation}
Here $w^{(t,k)}$ is a weight for combining the class probabilities of the
$t$-th tree from the $k$-th forest. An illustration of the weighted averaging
is shown in Fig. \ref{fig:weighted_class}, where we partly modify a picture
from \cite{Zhou-Feng-2017} (the left part is copied from \cite[Fig.
2]{Zhou-Feng-2017}) in order to show how elements of the class vector are
derived as a simple weighted sum. One can see from Fig.
\ref{fig:weighted_class} that three-class distribution is estimated by
counting the percentage of different classes of a new training example
$\mathbf{x}_{i}$ at the leaf node where the concerned example $\mathbf{x}_{i}$
falls into. Then the class vector of $\mathbf{x}_{i}$ is computed as the
weighted average. It is important to note that we weigh trees belonging to one
of the forests, but we do not weigh classes, i.e., the weights do not depend
on the class $c$. Moreover, the weights characterize trees, but not training
elements. One can also see from Fig. \ref{fig:weighted_class} that the
augmented features $v_{i,c}^{(k)}$, $c=1,...,C$, corresponding to the $q$-th
forest are obtained as weighted sums, i.e., there hold
\begin{align*}
v_{i,1}^{(k)}  &  =0.4w^{(1,k)}+0.2w^{(2,k)}+1.0w^{(3,k)},\\
v_{i,2}^{(k)}  &  =0.4w^{(1,k)}+0.5w^{(2,k)}+0.0w^{(3,k)},\\
v_{i,3}^{(k)}  &  =0.2w^{(1,k)}+0.3w^{(2,k)}+0.0w^{(3,k)}.
\end{align*}

The weights are restricted by the following obvious condition:
\begin{equation}
\sum_{t=1}^{T_{k}}w^{(t,k)}=1,\ w^{(t,k)}\geq0,\ t=1,...,T_{k}.
\label{DisDF_42}%
\end{equation}

So, we have the weighted averages for every forest, and the weights are
trained parameters which are optimized in order to decrease the distance
between objects from the same class and to increase the distance between
objects from different classes. Therefore, the next task is to develop an
algorithm for training the DisDF, in particular, for computing the weights for
every forest and for every cascade level.%

%TCIMACRO{\FRAME{ftbpFU}{5.2041in}{1.3602in}{0pt}{\Qcb{An illustration of the
%class vector generation taking into account the weights}}%
%{\Qlb{fig:weighted_class}}{weighted_class_vector_gen_3.png}%
%{\special{ language "Scientific Word";  type "GRAPHIC";
%maintain-aspect-ratio TRUE;  display "USEDEF";  valid_file "F";
%width 5.2041in;  height 1.3602in;  depth 0pt;  original-width 9.1316in;
%original-height 2.3645in;  cropleft "0";  croptop "1";  cropright "1";
%cropbottom "0";
%filename 'Weighted_Class_vector_gen_3.png';file-properties "XNPEU";}} }%
%BeginExpansion
%EndExpansion

\section{The DisDF training and testing}

In this section, we consider how to efficiently compute the weights in the
DisDF. The main difficulty in solving this task is to choose a proper loss
function which allows us to implement efficient computations. We overcome the
difficulty by modifying the well-known contrastive loss function. The proposed
modification gives us a convex loss function with respect to weights.

Before considering the DisDF training, we introduce the following notation:
\begin{align*}
P_{ij}^{(t,k)}  &  =\sum_{c=1}^{C}\left(  p_{i,c}^{(t,k)}-p_{j,c}%
^{(t,k)}\right)  ^{2},\\
Q_{ij}^{(t,k)}  &  =\sum_{c=1}^{C}\left \vert p_{i,c}^{(t,k)}-p_{j,c}%
^{(t,k)}\right \vert ,\\
\pi^{(k,t)}  &  =\sum_{i,j}(1-z_{ij})P_{ij}^{(t,k)},
\end{align*}%
\begin{align*}
\mathbf{P}_{ij}^{(k)}  &  =\left(  P_{ij}^{(t,k)},\ t=1,...,T_{k}\right)
,\  \mathbf{P}_{ij}=\left(  \mathbf{P}_{ij}^{(k)},\ k=1,...,M\right)  ,\\
\mathbf{Q}_{ij}^{(k)}  &  =\left(  Q_{ij}^{(t,k)},\ t=1,...,T_{k}\right)
,\  \mathbf{Q}_{ij}=\left(  \mathbf{Q}_{ij}^{(k)},\ k=1,...,M\right)  ,\\
\mathbf{w}^{(k)}  &  =\left(  w^{(t,k)},\ t=1,...,T_{k}\right)  ,\  \mathbf{w}%
=\left(  \mathbf{w}^{(k)},\ k=1,...,M\right)  ,\\
\left(  \mathbf{w}^{(k)}\right)  ^{2}  &  =\left(  \left(  w^{(t,k)}\right)
^{2},\ t=1,...,T_{k}\right)  ,\mathbf{w}^{2}=\left(  \left(  \mathbf{w}%
^{(k)}\right)  ^{2},\ k=1,...,M\right)  ,\\
\mathbf{\pi}^{(k)}  &  =\left(  \pi^{(t,k)},\ t=1,...,T_{k}\right)
,\  \mathbf{\pi}=\left(  \mathbf{\pi}^{(k)},\ k=1,...,M\right)  .
\end{align*}

Vectors $\mathbf{P}_{ij}$, $\mathbf{Q}_{ij}$, $\mathbf{w}$, $\mathbf{w}^{2}$,
$\mathbf{\pi}$ have the same length $\sum_{k=1}^{M}T_{k}$, and they are
produced as the concatenation of $M$ vectors characterizing the forests, for
example, the vector $\mathbf{w}$ is the concatenation of vectors
$\mathbf{w}^{(k)}=(w^{(1,k)},...,w^{(T_{k},k)})$, $k=1,...,M$. If a pair of
examples $(\mathbf{x}_{i},\mathbf{x}_{j})$ belongs to the same class, then we
assume that $z_{ij}=0$, otherwise $z_{ij}=1$.

We apply the greedy algorithm for training the DisDF, namely, we train
separately every level starting from the first level such that every next
level uses results of training at the previous level. The training process at
every level consists of two steps. The first step aims to train all trees by
applying all training examples. This step totally coincides with the training
algorithm of the original gcForest proposed by Zhou and Feng
\cite{Zhou-Feng-2017}.

The second step is to train the DisDF in order to get the weights $w^{(t,k)}$,
$t=1,...,T_{k}$. This can be done by minimizing the following objective
function over $M$ unit (probability) simplices in $\mathbb{R}^{T_{k}}$ denoted
as $\Delta_{k}$, i.e., over non-negative vectors $\mathbf{w}^{(k)}$,
$k=1,...,M$, that sum up to one:%
\begin{equation}
\min_{\mathbf{w}}J_{q}(\mathbf{w})=\min_{\mathbf{w}}\sum \nolimits_{i,j}%
l(\mathbf{x}_{i},\mathbf{x}_{j},y_{i},y_{j},\mathbf{w})+\lambda R(\mathbf{w}).
\label{DisDF_50}%
\end{equation}
Here $l$ is the loss function, $R(\mathbf{w})$ is a regularization term,
$\lambda$ is a hyper-parameter which controls the strength of the
regularization. We define the regularization term as
\[
R(\mathbf{w})=\left \Vert \mathbf{w}\right \Vert ^{2}.
\]

One of the most popular loss functions in metric learning is the contrastive
loss which can be written in terms of the considered problem as follows:%
\begin{equation}
l(\mathbf{x}_{i},\mathbf{x}_{j},y_{i},y_{j},\mathbf{w})=\left(  (1-z_{ij}%
)d\left(  \mathbf{x}_{i},\mathbf{x}_{j}\right)  +z_{ij}\max \left(
0,\tau-d\left(  \mathbf{x}_{i},\mathbf{x}_{j}\right)  \right)  \right)  .
\label{DisDF_52}%
\end{equation}

Here $\tau$ is the tuning parameter. It can be seen from the above expression
that the first term of the sum corresponds to the reduction of distances
between points of the same class, and the second term increases the distances
between points from different classes. We aim to find the minimum of the
function with respect to $\mathbf{w}$ in order to find an optimal vector of
weights which fulfills the above properties of examples from the same and from
different classes. Moreover, we aim to get the convex function $J_{q}%
(\mathbf{w})$ with respect to $\mathbf{w}$ satisfying (\ref{DisDF_42}) in
order to simplify the solution of the optimization problem.

Since the weights $\mathbf{w}$ impact on values of augmented parts of vectors
$\mathbf{x}_{i}$, then we define the distances between these parts of vectors
without taking into account the concatenated original vectors. Let us denote
the distance between two vectors $\mathbf{v}_{i}^{(k)}$ and $\mathbf{v}%
_{j}^{(k)}$ as $d\left(  \mathbf{v}_{i}^{(k)},\mathbf{v}_{j}^{(k)}\right)  $.
Then the $k$-th forest at level $q$ produces the following Euclidean
distance:
\begin{align*}
d\left(  \mathbf{v}_{i}^{(k)},\mathbf{v}_{j}^{(k)}\right)   &  =\sum_{c=1}%
^{C}\left(  v_{i,c}(k)-v_{j,c}(k)\right)  ^{2}\\
&  =\sum_{c=1}^{C}\sum_{t=1}^{T_{k}}\left(  p_{i,c}^{(t,k)}-p_{j,c}%
^{(t,k)}\right)  ^{2}\left(  w^{(t,k)}\right)  ^{2}\\
&  =\sum_{t=1}^{T_{k}}P_{ij}^{(t,k)}\left(  w^{(t,k)}\right)  ^{2}.
\end{align*}

Taking into account all $M$ forests at level $q$, we can write the total
distance between $\mathbf{v}_{i}$ and $\mathbf{v}_{j}$ without original
vectors as:%
\begin{equation}
d\left(  \mathbf{v}_{i},\mathbf{v}_{j}\right)  =\sum_{k=1}^{M}\sum
_{t=1}^{T_{k}}P_{ij}^{(t,k)}\left(  w^{(t,k)}\right)  ^{2}=\left \langle
\mathbf{P}_{ij},\mathbf{w}^{2}\right \rangle . \label{DisDF_58}%
\end{equation}

Here $<\cdot,\cdot>$ is the dot product of two vectors. The function
(\ref{DisDF_58}) is convex with respect to $w^{(t,k)}$. Unfortunately, the
second term in (\ref{DisDF_52}) as a function of $\mathbf{w}$ is non-convex.
In order to overcome this difficulty, we modify the contrastive loss and
reformulate the distance metric in the second term as follows:%
\[
d_{1}\left(  \mathbf{v}_{i},\mathbf{v}_{j}\right)  =\left \Vert \mathbf{v}%
_{i}-\mathbf{v}_{j}\right \Vert _{1}=\sum_{k=1}^{M}\sum_{t=1}^{T_{k}}%
Q_{ij}^{(t,k)}w^{(t,k)}=\left \langle \mathbf{Q}_{ij},\mathbf{w}\right \rangle
.
\]

The Manhattan distance is used instead of the Euclidean distance. We do not
need to consider the absolute values of $w^{(t,k)}$ because the weights are
restricted by condition (\ref{DisDF_42}) such that their values are
non-negative. Moreover, we rewrite the second term in the following form:%
\[
z_{ij}\left(  \max \left(  0,\tau-d_{1}\left(  \mathbf{v}_{i},\mathbf{v}%
_{j}\right)  \right)  \right)  ^{2}.
\]

The above function is convex in the interval $[0,1]$ of $w^{(t,k)}$. Then the
objective function $J_{q}(\mathbf{w})$ as a sum of the convex functions is
convex with respect to weights. Finally, we write the function $J_{q}%
(\mathbf{w})$ as
\begin{align}
J_{q}(\mathbf{w})  &  =\min_{\mathbf{w}}\sum_{i,j}(1-z_{ij})\left \langle
\mathbf{P}_{ij},\mathbf{w}^{2}\right \rangle \nonumber \\
&  +\sum_{i,j}z_{ij}\left(  \max \left(  0,\tau-\left \langle \mathbf{Q}%
_{ij},\mathbf{w}\right \rangle \right)  \right)  ^{2}+\lambda \left \Vert
\mathbf{w}\right \Vert ^{2}. \label{DisDF_62}%
\end{align}

Let us introduce new variables
\[
\alpha_{ij}=\max \left(  0,\tau-\left \langle \mathbf{Q}_{ij},\mathbf{w}%
\right \rangle \right)  .
\]
Then we can rewrite the optimization problem as
\begin{equation}
J_{q}(\mathbf{w})=\min_{\alpha_{ij},\mathbf{w}}\left(  \left \langle
\mathbf{\pi},\mathbf{w}^{2}\right \rangle +\sum_{i,j}z_{ij}\alpha_{ij}%
^{2}+\lambda \left \Vert \mathbf{w}\right \Vert ^{2}\right)  , \label{DisDF_64}%
\end{equation}
subject to (\ref{DisDF_42}) and
\begin{equation}
\alpha_{ij}\geq \tau-\left \langle \mathbf{Q}_{ij},\mathbf{w}\right \rangle
,\  \  \alpha_{ij}\geq0,\  \  \forall i,j. \label{DisDF_66}%
\end{equation}

This is a standard quadratic optimization problem with linear constraints and
with $M\cdot T_{k}$ variables $w^{(t,k)}$ and $n\cdot(n-1)/2$ variables
$\alpha_{ij}$.

Let us prove that the problem can be decomposed into $M$ quadratic
optimization problems in accordance with forests, i.e., every optimization
problem can be solved for every forest separately. Let us return to the
problem (\ref{DisDF_62}). The first term in the objective function can be
rewritten as
\[
\sum_{i,j}(1-z_{ij})\left \langle \mathbf{P}_{ij},\mathbf{w}^{2}\right \rangle
=\sum_{k=1}^{M}\left[  \sum_{i,j}(1-z_{ij})\sum_{t=1}^{T_{k}}P_{ij}%
^{(t,k)}\left(  w^{(t,k)}\right)  ^{2}\right]  .
\]
Since constraints (\ref{DisDF_42}) for $w^{(t,k_{0})}$ and $w^{(t,k_{1})}$ do
not intersect each other by $k_{0}\neq k_{1}$, i.e., they consist of different
weights, then we can separately consider $M$ optimization problems for every
$k$.

Let us consider the second term in (\ref{DisDF_62}) and rewrite it as
follows:
\begin{align*}
&  \sum_{i,j}z_{ij}\left(  \max \left(  0,\tau-\left \langle \mathbf{Q}%
_{ij},\mathbf{w}\right \rangle \right)  \right)  ^{2}\\
&  =\sum_{i,j}z_{ij}\left(  \max \left(  0,\sum_{k=1}^{M}\left[  \tau_{k}%
-\sum_{t=1}^{T_{k}}Q_{ij}^{(t,k)}\left(  w^{(t,k)}\right)  \right]  \right)
\right)  ^{2}.
\end{align*}

Here $\sum_{k=1}^{M}\tau_{k}=\tau$. In order to minimize $J_{q}(\mathbf{w})$,
we need to maximize $\tau_{k}-\sum_{t=1}^{T_{k}}Q_{ij}^{(t,k)}\left(
w^{(t,k)}\right)  $. But its maximizing does not depend on $k$ because the
corresponding constraints (\ref{DisDF_42}) for $w^{(t,k_{0})}$ and
$w^{(t,k_{1})}$ do not intersect each other. Since $\tau$ as well as $\tau
_{k}$ are tuning parameters, then we can tune $\tau_{k}$ for every
$k=1,...,M$, separately. Then we can write
\[
\sum_{i,j}z_{ij}\left(  \max \left(  0,\left[  \tau_{k_{0}}+A-\sum_{t=1}%
^{T_{k}}Q_{ij}^{(t,k)}w^{(t,k)}\right]  \right)  \right)  ^{2}.
\]
Here $A$ is a term which does not depend on $w^{(t,k)}$, but, of course, it
depends on other weights from the vector $\mathbf{w}^{(k)}$. In fact, we have
the parameter $\tau_{k}=\tau_{k_{0}}+A$ for tuning. So, we can separately
solve the problems for every forest. Then (\ref{DisDF_64})-(\ref{DisDF_66})
can be rewritten for every forest as follows:%
\begin{equation}
J_{q}(\mathbf{w}^{(k)})=\min_{\alpha_{ij},\mathbf{w}^{(k)}}\left(
\left \langle \mathbf{\pi}^{(k)},\left(  \mathbf{w}^{(k)}\right)
^{2}\right \rangle +\sum_{i,j}z_{ij}\alpha_{ij}^{2}+\lambda \left \Vert
\mathbf{w}^{(k)}\right \Vert ^{2}\right)  , \label{DisDF_70}%
\end{equation}
subject to (\ref{DisDF_42}) and
\begin{equation}
\alpha_{ij}\geq \tau-\left \langle \mathbf{Q}_{ij}^{(k)},\mathbf{w}%
^{(k)}\right \rangle ,\  \  \alpha_{ij}\geq0,\  \  \forall i,j. \label{DisDF_72}%
\end{equation}

In sum, we can write a general algorithm for training the DisDF (see Algorithm
\ref{alg:DisDF_4}). Its complexity mainly depends on the number of levels.
Having the trained DisDF, we can classify a new example $\mathbf{x}$. By using
the trained decision trees and the weights $\mathbf{w}$, the vector
$\mathbf{x}$ is augmented at each level. Finally, we get the vector
$\mathbf{v}$ of augmented features after the $Q$-th level of the forest
cascade corresponding to original example $\mathbf{x}$.

The class of the example $\mathbf{x}$ is defined by the sum of the $c$-th
elements of vectors $\mathbf{v}^{(1,Q)},...,\mathbf{v}^{(M_{Q},Q)}$. The
example $\mathbf{x}$ belongs to the class $c$, if the sum of the $c$-th
elements $v_{c}^{(1,Q)}+...+v_{c}^{(M_{Q},Q)}$ is maximal.

\begin{algorithm}
\caption{A general algorithm for training the DDF} \label{alg:DisDF_4}

\begin{algorithmic}
[1]\REQUIRE Training set $S$ $=\{(\mathbf{x}_{i},y_{i}),\ i=1,...,n\}$,
$\mathbf{x}_{i}\in \mathbb{R}^{m}$, $y_{i}\in \{1,...,C\}$; number of levels $Q$

\ENSURE$\mathbf{w}$ for every $q=1,...,Q$

\FOR{$q=1$, $q\leq Q$ } \STATE Train all trees at the $q$-th level in
accordance with the gcForest algorithm \cite{Zhou-Feng-2017}

\FOR{$k=1$, $k\leq M_q$ } \STATE Compute the weights $\mathbf{w}^{(k)}$ by
solving the $k$-th quadratic optimization problem with the objective function
(\ref{DisDF_70}) and constraints (\ref{DisDF_42}) and (\ref{DisDF_72})

\ENDFOR

\STATE Concatenate $\mathbf{w}^{(k)}$, $k=1,...,M_{q}$, to get $\mathbf{w}$

\STATE For every $\mathbf{x}_{i}$, compute $\mathbf{v}_{i}^{(k)}$ by using
(\ref{DisDF_41}), $k=1,...,M_{q}$, and $\mathbf{v}_{i}$

\STATE For every $\mathbf{x}_{i}$, form the concatenated vector $\mathbf{x}%
_{i}^{(q)}$ for the next level by using (\ref{DisDF_40})

\ENDFOR

\end{algorithmic}
\end{algorithm}

\section{Large-scale data}

The main difficulty in solving the problem (\ref{DisDF_70})-(\ref{DisDF_72})
is that the number of variables as well as the number of constraints may be
extremely large because we need to enumerate all pairs of training data.
Therefore, we simplify the algorithm by using the well-known Frank-Wolfe
algorithm \cite{Frank-Wolfe-1956}. It is represented as Algorithm
\ref{alg:DisDF_6}.

\begin{algorithm}
\caption{The Frank-Wolfe algorithm \cite{Frank-Wolfe-1956} in terms of the DDF training}
\label{alg:DisDF_6}

\begin{algorithmic}
[1]\REQUIRE$\mathbf{P}_{ij}^{(k)}$, $\mathbf{Q}_{ij}^{(k)}$, $z_{ij}$,
$\lambda$, $\tau$; number of iterations $S$

\ENSURE$\mathbf{w}^{(k)}$

\STATE Initialize $\mathbf{w}_{0}\in \Delta_{k}$, $\Delta_{k}$ is the unit
simplex having $T_{k}$ vertices

\FOR{$s=0$, $s\leq S-1$ } \STATE Compute the gradient $f(\mathbf{w}%
_{s})=\nabla_{\mathbf{w}^{(k)}}J_{q}(\mathbf{w}_{s})$

\STATE Solve the linear problem $\mathbf{g}_{s}\leftarrow \arg \min
_{\mathbf{g}\in \Delta_{k}}\left \langle \mathbf{g},f(\mathbf{w}_{s}%
)\right \rangle $, $\mathbf{g\in}\mathbb{R}^{T_{k}}$ is the vector of
optimization variables

\STATE Compute $\gamma_{s}\leftarrow2/(s+2)$

\STATE Update $\mathbf{w}_{s+1}\leftarrow \mathbf{w}_{s}+\gamma_{s}\left(
\mathbf{g}_{s}-\mathbf{w}_{s}\right)  $ \ENDFOR

\STATE$\mathbf{w}^{(k)}\leftarrow \mathbf{w}_{s+1}$
\end{algorithmic}
\end{algorithm}

Denote
\[
S_{ij}=\tau-\left \langle \mathbf{Q}_{ij}^{(k)},\mathbf{w}^{(k)}\right \rangle
.
\]
The gradient of the function $J_{q}(\mathbf{w}^{(k)})$ with respect to the
variable $w^{(k,t)}$ is
\begin{align}
\nabla_{w^{(k,t)}}J_{q}(\mathbf{w}^{(k)})  &  =2w^{(k,t)}\left[  \lambda
+\sum_{i,j}(1-z_{ij})P_{ij}^{(k,t)}\right] \nonumber \\
&  -2\sum_{i,j}z_{ij}\cdot \left \{
\begin{array}
[c]{cc}%
0, & S_{ij}\leq0,\\
S_{ij}^{2}\cdot Q_{ij}^{(k,t)}, & S_{ij}>0.
\end{array}
\right.  \label{DisDF_78}%
\end{align}

It should be noted that the linear problem in Algorithm \ref{alg:DisDF_6} can
be solved by looking for the solution among $T_{k}$ vertices of the unit
simplex $\Delta_{k}$. The vertices are of the form:
\[
\mathbf{g=}(0,...,0,1,0,...,0).
\]
Hence, we have to look for smallest value of $f(\mathbf{w}_{s})$ by
$t=1,...,T_{k}$. In other words, we compute $\nabla_{\mathbf{w}^{(k)}}%
J_{q}(\mathbf{w}^{(k)})$ for different $t$. Then $\mathbf{g}_{s}$ consists of
$T_{k}-1$ zero elements and the unit element whose index $t_{0}$ coincides
with the index of the smallest value of $f(\mathbf{w}_{s})$.

So, we have obtained a very simple algorithm for solving the problem
(\ref{DisDF_70})-(\ref{DisDF_72}). Its simplicity is due to simple constraints
for the weights $w^{(k,t)}$. The modified Frank-Wolfe algorithm is represented
as Algorithm \ref{alg:DisDF_8}.

\begin{algorithm}
\caption{The modified Frank-Wolfe algorithm} \label{alg:DisDF_8}

\begin{algorithmic}
[1]\REQUIRE$\mathbf{P}_{ij}^{(k)}$, $\mathbf{Q}_{ij}^{(k)}$, $z_{ij}$,
$\lambda$, $\tau$; number of iterations $S$

\ENSURE$\mathbf{w}^{(k)}$

\STATE Initialize $\mathbf{w}_{0}\in \Delta_{k}$, $\Delta_{k}$ is the unit
simplex having $T_{k}$ vertices

\FOR{$s=0$, $s\leq S-1$ }

\STATE Compute $\nabla_{w^{(k,t)}}J_{q}(\mathbf{w}^{(k)})$ for every
$t=1,...,T_{k}$ by using (\ref{DisDF_78})

\STATE Compute $t_{0}\leftarrow \arg \min_{t=1,...,T_{k}}\nabla_{w^{(k,t)}}%
J_{q}(\mathbf{w}^{(k)})$

\STATE$\mathbf{g}_{s}\leftarrow(0,...,0,1_{t_{0}},0,...,0)$

\STATE Compute $\gamma_{s}\leftarrow2/(s+2)$

\STATE Update $\mathbf{w}_{s+1}\leftarrow \mathbf{w}_{s}+\gamma_{s}\left(
\mathbf{g}_{s}-\mathbf{w}_{s}\right)  $ \ENDFOR

\STATE$\mathbf{w}^{(k)}\leftarrow \mathbf{w}_{s+1}$
\end{algorithmic}
\end{algorithm}

\section{Numerical experiments}

We compare the DisDF with the gcForest. The DisDF has the same cascade
structure as the standard gcForest described in \cite{Zhou-Feng-2017}. Each
level (layer) of the cascade structure consists of 2 complete-random tree
forests and 2 random forests. Three-fold cross-validation is used for the
class vector generation. The number of cascade levels is automatically determined.

We modify a software in Python implementing the gcForest and available at
https://github.com/leopiney/deep-forest to implement the procedure for
computing optimal weights and weighted averages $v_{i,c}^{(k)}$. Accuracy
measure $A$ used in numerical experiments is the proportion of correctly
classified cases on a sample of data. To evaluate the average accuracy, we
perform a cross-validation with $100$ repetitions, where in each run, we
randomly select $N$ training data and $N_{\text{test}}=2N/3$ test data.

First, we compare the DisDF with the gcForest by using some public data sets
from UCI Machine Learning Repository \cite{Lichman:2013}: the Ecoli data set
(336 instances, 8 features, 8 classes), the Parkinsons data set (197
instances, 23 features, 2 classes), the Ionosphere data set (351 instances, 34
features, 2 classes). A more detailed information about the data sets can be
found from, respectively, the data resources. Different values for the
regularization hyper-parameter $\lambda$ have been tested, choosing those
leading to the best results. In order to investigate how the number of
decision trees impact on the classification accuracy, we study the DisDF as
well as gcForest by different number of trees, namely, we take $T_{k}=T=100$,
$400$, $700$, $1000$.

Results of numerical experiments for the Parkinsons data set are shown in
Table \ref{t:DisDF_4}. It contains the DisDF accuracy measures obtained for
the gcForest (denoted as gcF) and the DisDF as functions of the number of
trees $T$ in every forest and the number $N=50,80,100,120$ of examples in the
training set. It follows from Table \ref{t:DisDF_4} that the accuracy of the
DisDF exceeds the same measure of the gcForest in most cases. The difference
is not significant by $N=50$ and $120$. However, it is larger by $N=100$ and
the small amount of trees $T$. Nevertheless, the largest difference between
accuracy measures of the DisDF and the gcForest is observed by $T=1000$ and
$N=100$.

Results of numerical experiments for the Ecoli data set are shown in Table
\ref{t:DisDF_5}. This data set shows the largest difference between accuracy
measures of the DisDF and gcForest by $N=50$. This implies that the proposed
DisDF outperforms the gcForest by the very small amount of training data. It
is interesting to note that the DisDF does not outperform the gcForest by
$N=120$ and $T=700$. At the same time, the number of trees also significantly
impact on the accuracy, namely, we can see from Table \ref{t:DisDF_5} that the
outperformance of the DisDF is observed by $T=100$ and $400$.

Numerical results for the Ionosphere data set are represented in Table
\ref{t:DisDF_7}. It follows from Table \ref{t:DisDF_7} that the largest
difference between accuracy measures of the DisDF and the gcForest is observed
by $T=100$ and $N=50$. This again implies that the DisDF outperforms the
gcForest by the very small amount of training data.

By analyzing all results, we have to point out that, in contrast to
comparative results, the highest accuracy measure can be obtained by the large
number of training data and the large number of trees in every forest. If the
first parameter ($N$) cannot be controlled, then the number of trees is a
tuning parameter. It is interesting to note that the largest number of trees
by some fixed values of $N$, for example, $N=80$ and $100$, does not provide
the largest accuracy measure. In particular, we can see from Tables
\ref{t:DisDF_5}-\ref{t:DisDF_7} that the largest accuracy measures are
achieved at $T=400$ and $700$.%

%TCIMACRO{\TeXButton{B}{\begin{table}[tbp] \centering}}%
%BeginExpansion
\begin{table}[tbp] \centering
%EndExpansion
\caption{The DisDF accuracy for the Parkinsons data set by different $N$ and $T$ in every forest}%
\begin{tabular}
[c]{ccccccccc}\hline
$T$ & \multicolumn{2}{c}{$100$} & \multicolumn{2}{c}{$400$} &
\multicolumn{2}{c}{$700$} & \multicolumn{2}{c}{$1000$}\\ \hline
$N$ & gcF & DisDF & gcF & DisDF & gcF & DisDF & gcF & DisDF\\ \hline
$50$ & $0.80$ & $0.84$ & $0.85$ & $0.87$ & $0.84$ & $0.84$ & $0.79$ &
$0.80$\\ \hline
$80$ & $0.815$ & $0.87$ & $0.875$ & $0.90$ & $0.85$ & $0.875$ & $0.80$ &
$0.85$\\ \hline
$100$ & $0.833$ & $0.86$ & $0.88$ & $0.92$ & $0.86$ & $0.88$ & $0.83$ &
$0.90$\\ \hline
$120$ & $0.84$ & $0.883$ & $0.92$ & $0.95$ & $0.92$ & $0.95$ & $0.92$ &
$0.95$\\ \hline
\end{tabular}
\label{t:DisDF_4}%
%TCIMACRO{\TeXButton{E}{\end{table}}}%
%BeginExpansion
\end{table}%
%EndExpansion
%

%TCIMACRO{\TeXButton{B}{\begin{table}[tbp] \centering}}%
%BeginExpansion
\begin{table}[tbp] \centering
%EndExpansion
\caption{The DisDF accuracy for the Ecoli data set by different $N$ and $T$ in every forest}%
\begin{tabular}
[c]{ccccccccc}\hline
$T$ & \multicolumn{2}{c}{$100$} & \multicolumn{2}{c}{$400$} &
\multicolumn{2}{c}{$700$} & \multicolumn{2}{c}{$1000$}\\ \hline
$N$ & gcF & DisDF & gcF & DisDF & gcF & DisDF & gcF & DisDF\\ \hline
$50$ & $0.78$ & $0.82$ & $0.80$ & $0.86$ & $0.72$ & $0.76$ & $0.73$ &
$0.77$\\ \hline
$80$ & $0.88$ & $0.89$ & $0.81$ & $0.83$ & $0.87$ & $0.90$ & $0.84$ &
$0.88$\\ \hline
$100$ & $0.85$ & $0.90$ & $0.85$ & $0.87$ & $0.82$ & $0.84$ & $0.90$ &
$0.93$\\ \hline
$120$ & $0.84$ & $0.88$ & $0.80$ & $0.85$ & $0.93$ & $0.92$ & $0.95$ &
$0.96$\\ \hline
\end{tabular}
\label{t:DisDF_5}%
%TCIMACRO{\TeXButton{E}{\end{table}}}%
%BeginExpansion
\end{table}%
%EndExpansion
%

%TCIMACRO{\TeXButton{B}{\begin{table}[tbp] \centering}}%
%BeginExpansion
\begin{table}[tbp] \centering
%EndExpansion
\caption{The DisDF accuracy for the Ionsphere data set by different $N$ and $T$ in every forest}%
\begin{tabular}
[c]{ccccccccc}\hline
$T$ & \multicolumn{2}{c}{$100$} & \multicolumn{2}{c}{$400$} &
\multicolumn{2}{c}{$700$} & \multicolumn{2}{c}{$1000$}\\ \hline
$N$ & gcF & DisDF & gcF & DisDF & gcF & DisDF & gcF & DisDF\\ \hline
$50$ & $0.48$ & $0.66$ & $0.58$ & $0.50$ & $0.48$ & $0.58$ & $0.40$ &
$0.60$\\ \hline
$80$ & $0.71$ & $0.78$ & $0.68$ & $0.70$ & $0.66$ & $0.65$ & $0.72$ &
$0.75$\\ \hline
$100$ & $0.72$ & $0.80$ & $0.74$ & $0.78$ & $0.74$ & $0.78$ & $0.76$ &
$0.775$\\ \hline
$120$ & $0.69$ & $0.70$ & $0.77$ & $0.80$ & $0.81$ & $0.82$ & $0.83$ &
$0.83$\\ \hline
\end{tabular}
\label{t:DisDF_7}%
%TCIMACRO{\TeXButton{E}{\end{table}}}%
%BeginExpansion
\end{table}%
%EndExpansion

It should be noted that the multi-grained scanning proposed in
\cite{Zhou-Feng-2017} was not applied to investigating the above data sets
having relatively small numbers of features. The above numerical results have
been obtained by using only the forest cascade structure.

Another data set for comparison of the DisDF and gcForest is the well-known
MNIST data set which is a commonly used large database of $28\times28$ pixel
handwritten digit images \cite{LeCun-etal-1998}. It has a training set of
60,000 examples, and a test set of 10,000 examples. The digits are
size-normalized and centered in a fixed-size image. The data set is available
at http://yann.lecun.com/exdb/mnist/. In contrast to gcForest, we did not use
the multi-grained scanning scheme for the DisDF implementation because its use
provides worse results by the small amount of training data. Results of
numerical experiments for the MNIST data set are shown in Table
\ref{t:DisDF_8}. Numerical experiments with the MNIST data set by the very
small training data have shown that the use of the multi-grained scanning
procedure may deteriorate the classification performance of the DisDF as well
as the gcForest. Therefore, we did not use this procedure in numerical
experiments with the MNIST. It can be seen from Table \ref{t:DisDF_8} that the
DisDF outperforms the gcForest in the most cases. The largest difference
between accuracy measures of the DisDF and the gcForest is observed by $T=100$
and $N=50$.%

%TCIMACRO{\TeXButton{B}{\begin{table}[tbp] \centering}}%
%BeginExpansion
\begin{table}[tbp] \centering
%EndExpansion
\caption{The DisDF accuracy for the MNIST data set by different $N$ and $T$ in every forest}%
\begin{tabular}
[c]{ccccccccc}\hline
$T$ & \multicolumn{2}{c}{$100$} & \multicolumn{2}{c}{$400$} &
\multicolumn{2}{c}{$700$} & \multicolumn{2}{c}{$1000$}\\ \hline
$N$ & gcF & DisDF & gcF & DisDF & gcF & DisDF & gcF & DisDF\\ \hline
$50$ & $0.63$ & $0.70$ & $0.67$ & $0.71$ & $0.68$ & $0.68$ & $0.69$ &
$0.70$\\ \hline
$80$ & $0.63$ & $0.70$ & $0.75$ & $0.77$ & $0.75$ & $0.77$ & $0.70$ &
$0.70$\\ \hline
$100$ & $0.70$ & $0.73$ & $0.76$ & $0.78$ & $0.76$ & $0.78$ & $0.76$ &
$0.78$\\ \hline
$120$ & $0.74$ & $0.75$ & $0.76$ & $0.76$ & $0.76$ & $0.78$ & $0.77$ &
$0.78$\\ \hline
\end{tabular}
\label{t:DisDF_8}%
%TCIMACRO{\TeXButton{E}{\end{table}}}%
%BeginExpansion
\end{table}%
%EndExpansion

\section{Conclusion}

A discriminative metric learning algorithm in the form of the DisDF has been
presented in the paper. Two main contributions should be pointed out. First,
we have introduced weights for trees which allow us to apply some new
properties for the deep forest. The weights play a key role in the developing
the DisDF. This role is similar to the role of weights of connections in
neural networks which also have to be trained. This implies that we can
control properties of the deep forest and its modifications by constructing
the corresponding objective functions $J(\mathbf{w})$ which are similar to the
loss or reconstruction functions in neural networks. This fact opens a way for
developing new modifications of the deep forest which have certain properties.
Moreover, we get an opportunity to consider the relationship between the deep
forest and neural networks as it has been done by Richmond et al.
\cite{Richmond-etal-2015} in exploring the relationship between stacked random
forests and deep convolutional neural networks. Second, we have used a new
objective function $J_{q}(\mathbf{w})$ which combines two different distance
metrics: Euclidean and Manhattan distances. This combination allows us to get
the convex objective function with respect to $\mathbf{w}$ and to
significantly simplify the optimization problem.

It should be also noted that one of the implementations of the DisDF has been
represented in the paper. It should be noted that other modifications of the
DisDF can be also obtained. We can use more efficient modifications of the
Frank-Wolfe algorithm, for example, algorithms proposed by Hazan and Luo
\cite{Hazan-Luo-2016} or Reddi et al. \cite{Reddi-etal-2016}. We can also
consider non-linear functions of weights like activation functions in neural
networks. Moreover, we can investigate imprecise statistical models
\cite{Walley91} for restricting the set of weights, for example, we can reduce
the unit simplex of weights in order to get robust classification models.
These modifications can be viewed as directions for further research.

\section*{Acknowledgement}

The reported study was partially supported by RFBR, research project No. 17-01-00118.

\end{document}